\documentclass[conference]{IEEEtran}
\IEEEoverridecommandlockouts
\usepackage{hyperref}
\usepackage{cite}
\usepackage{enumitem}
\usepackage{amsmath,amssymb,amsfonts}
\usepackage{algorithmic}
\usepackage{graphicx}
\usepackage{textcomp}
\usepackage{bm}  
\usepackage{multirow}
\usepackage{float}
\usepackage{xcolor}
\usepackage{booktabs}
\usepackage{pifont}     
\usepackage{amssymb}    
\usepackage{multirow}
\usepackage{array}
\def\BibTeX{{\rm B\kern-.05em{\sc i\kern-.025em b}\kern-.08em
    T\kern-.1667em\lower.7ex\hbox{E}\kern-.125emX}}
\begin{document}

\title{RTGMFF: Enhanced fMRI-based Brain Disorder Diagnosis via ROI-driven Text Generation and Multimodal Feature Fusion\\

}

\author{%
\IEEEauthorblockN{1\textsuperscript{st} Junhao Jia}
\IEEEauthorblockA{\textit{Hangzhou Dianzi University}\\
Hangzhou, China\\
23080631@hdu.edu.cn}
\and
\IEEEauthorblockN{2\textsuperscript{nd} Yifei Sun}
\IEEEauthorblockA{\textit{Zhejiang University}\\
Hangzhou, China\\
diaoquesang@gmail.com}
\and
\IEEEauthorblockN{3\textsuperscript{rd} Yunyou Liu}
\IEEEauthorblockA{\textit{Hangzhou Dianzi University}\\
Hangzhou, China\\
23150118@hdu.edu.cn}
\and
\IEEEauthorblockN{4\textsuperscript{th} Cheng Yang}
\IEEEauthorblockA{\textit{Hangzhou Dianzi University}\\
Hangzhou, China\\
23320214@hdu.edu.cn}
\and
\IEEEauthorblockN{5\textsuperscript{th} Changmiao Wang*}
\IEEEauthorblockA{\textit{Shenzhen Research Institute of Big Data}\\
Shenzhen, China\\
cmwangalbert@gmail.com}
\and
\IEEEauthorblockN{6\textsuperscript{th} Feiwei Qin*}
\IEEEauthorblockA{\textit{Hangzhou Dianzi University}\\
Hangzhou, China\\
qinfeiwei@hdu.edu.cn}
\and
\IEEEauthorblockN{7\textsuperscript{th} Yong Peng}
\IEEEauthorblockA{\textit{Hangzhou Dianzi University}\\
Hangzhou, China\\
yongpeng@hdu.edu.cn}
\and
\IEEEauthorblockN{8\textsuperscript{th} Wenwen Min}
\IEEEauthorblockA{\textit{Yunnan University}\\
Kunming, China\\
minwenwen@ynu.edu.cn}
}

\maketitle

\begin{abstract}
Functional magnetic resonance imaging (fMRI) is a powerful tool for probing brain function, yet reliable clinical diagnosis is hampered by low signal-to-noise ratios, inter-subject variability, and the limited frequency awareness of prevailing CNN- and Transformer-based models. Moreover, most fMRI datasets lack textual annotations that could contextualize regional activation and connectivity patterns. We introduce RTGMFF, a framework that unifies automatic ROI-level text generation with multimodal feature fusion for brain-disorder diagnosis. RTGMFF consists of three components: (i) ROI-driven fMRI text generation deterministically condenses each subject's activation, connectivity, age, and sex into reproducible text tokens; (ii) Hybrid frequency-spatial encoder fuses a hierarchical wavelet-mamba branch with a cross-scale Transformer encoder to capture frequency-domain structure alongside long-range spatial dependencies; and (iii) Adaptive semantic alignment module embeds the ROI token sequence and visual features in a shared space, using a regularized cosine-similarity loss to narrow the modality gap. Extensive experiments on the ADHD-200 and ABIDE benchmarks show that RTGMFF surpasses current methods in diagnostic accuracy, achieving notable gains in sensitivity, specificity, and area under the ROC curve. Code is available at \href{https://github.com/BeistMedAI/RTGMFF}{https://github.com/BeistMedAI/RTGMFF}.
\end{abstract}


\section{Introduction}
The rapid advancement of interdisciplinary research in neuroimaging and computer vision has positioned functional magnetic resonance imaging (fMRI) as a vital tool for studying dynamic brain activity~\cite{klohs2025advanced}. Its applications in neuroscience and clinical diagnosis are expanding, offering significant potential for understanding brain function \cite{yang2025applications}. However, the analysis of fMRI data comes with substantial challenges due to its high dimensionality, inherent noise, and pronounced individual variability. These complexities hinder the precise identification of both localized brain region activations and the intricate functional connectivity between regions. Further complicating analysis is the influence of demographic factors such as age on brain function \cite{reiss1996brain}. Traditional methods often struggle to integrate these variables while simultaneously addressing both local and global brain network characteristics. This limitation highlights the ongoing challenge of extracting meaningful and clinically actionable insights from multimodal fMRI data.

Recent advancements in multi-scale feature extraction and fusion have significantly improved fMRI data analysis. CNN-based models~\cite{hornik1989multilayer, he2016deep, simonyan2014very, kawahara2017brainnetcnn, jia2025mmafn} are effective at capturing local spatial features but struggle with long-range dependencies. Transformer-based models~\cite{liu2021swin, kan2022brain, bannadabhavi2023community, zhang2023multi, jiang2024multi, peng2024gbt, shi2025transformer, shehzad2025dynamic}, while theoretically well-suited for modeling global interactions, have been successfully applied to fMRI-based brain disorder diagnosis tasks such as ADHD classification~\cite{shi2025transformer} and ASD detection in ABIDE~\cite{shehzad2025dynamic}, but are often used in ways that focus primarily on spatial relationships, with limited attention to frequency-domain representations and temporal dynamics, both of which are critical for understanding brain function.
In contrast, emerging architectures such as Mamba \cite{gu2023mamba} and wavelet-based methods \cite{cohen2019better} offer promising approaches for integrating frequency and spatial domain features through multi-scale analysis. Early progress \cite{zhang2024cf} has been made in areas like cross-scale feature fusion, semantic alignment, and multimodal collaborative modeling, demonstrating the potential of these strategies. Despite these advances, two critical challenges remain. First, fMRI data typically lack accompanying textual descriptions, which are vital for interpreting brain-region connectivity and activity.  Second, existing feature-extraction models underuse frequency-domain cues and rarely integrate frequency-spatial information within a unified representation.

To address these limitations, we propose RTGMFF, an enhanced fMRI-based brain disorder diagnosis pipeline that leverages ROI-driven fMRI text generation and multimodal feature fusion. Our contributions are summarized as follows: 
\begin{enumerate}[label=(\roman*), leftmargin=1.5em, itemsep=0.3em, parsep=0em]
    \item A deterministic ROI-driven fMRI text generator that fuses subject-specific activation statistics with demographic information to yield compact, reproducible tokens.
    
    \item A hybrid frequency-spatial encoder that couples a hierarchical wavelet–mamba branch with a cross-scale Transformer encoder, jointly modeling frequency-domain structure and long-range spatial dependencies.
    
    \item An adaptive semantic alignment module that embeds the ROI token sequence and aligns it with visual features through learnable projections and a cosine-based alignment loss with regularization to mitigate modality mismatch.
\end{enumerate}

\begin{figure*}[t]
\centering
\includegraphics[width=0.95\textwidth]{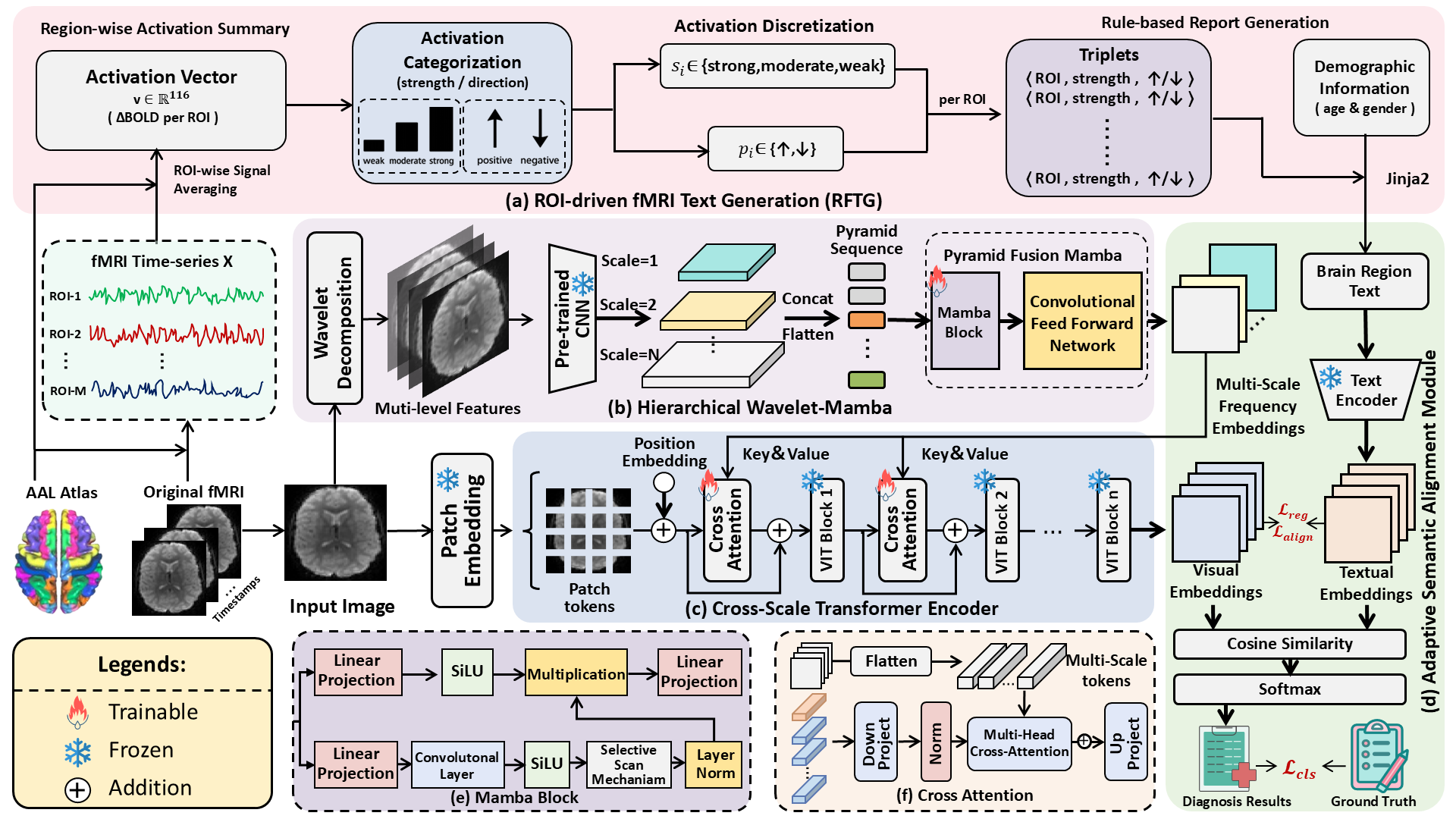}
\caption{The overview of our proposed RTGMFF pipeline. Panel (a) illustrates the ROI-driven fMRI Text Generation. Panel (b) is a Hierarchical Wavelet-Mamba architecture, and panel (c) illustrates the Cross-Scale Transformer Encoder, with the details of the Mamba Block and Cross Attention in panels (e) and (f). The network structure of the Adaptive Semantic Alignment Module is given in Panel (d).}
\label{stta}
\end{figure*}

\section{Related Work}

\subsection{fMRI-based Brain Disorder Classification}
Early efforts in fMRI-based diagnosis utilized universal vision models, including MLPs, VGG, and ResNet, as foundational frameworks~\cite{hornik1989multilayer, simonyan2014very, he2016deep}. These approaches evolved to incorporate hierarchical and full-attention Transformers, which are more effective at capturing long-range spatial dependencies~\cite{liu2021swin, dosovitskiy2020image}. Subsequent advancements in neuroimaging-specific models have further enhanced both the performance and interpretability of these diagnostic tools. These models harness the connectomic structure of the brain~\cite{jia2025brain}, exemplified by BrainNetCNN, which focuses on edge-to-node operations~\cite{kawahara2017brainnetcnn}, and graph-centric methods like BrainGNN and BrainGB~\cite{li2021braingnn, cui2022braingb}. Recent work has also explored connectome-level Transformer architectures that integrate both structural and functional modalities for ASD diagnosis with an emphasis on model interpretability~\cite{nazari2025explainable}. Additionally, community- and curriculum-aware Transformers have been developed to refine diagnostic accuracy~\cite{kan2022brain, bannadabhavi2023community, jiang2024multi, peng2024gbt}. These models enhance robustness against variations between different research sites through techniques like contrastive learning and domain adaptation~\cite{zhang2023gcl, xu2023cardiac, fang2025source, zeng2025knowledge}. Despite these technological strides, two significant gaps remain. First, frequency-domain information is often overlooked, even though resting-state BOLD data is characterized by distinct low-frequency patterns. Second, there is a lack of a systematic approach to effectively integrate both local and global representations.

\subsection{Multi-Scale and Frequency-Domain Representations}
Multi-scale representations are well motivated for neuroimaging due to the nested organization of functional systems. While CNN pyramids and Transformer hierarchies implicitly encode multi-scale features, explicit spectral modeling can provide complementary robustness and noise suppression~\cite{wang2024computing,jia2025geodesic}. Wavelet-based analysis enables localized time–frequency decomposition and has long been advocated for neuroimaging signals~\cite{cohen2019better}. Beyond classical discrete wavelets, scattering transforms offer stable, deformation-insensitive multi-scale features that preserve high-frequency content through cascaded wavelet moduli and averaging~\cite{bruna2013invariant}. However, many pipelines either use wavelets as shallow pre-processing or do not tightly couple spectral cues with downstream sequence modeling. In contrast, our framework performs multi-scale wavelet decomposition followed by selective scanning and enhancement, allowing frequency-aware locality to interact directly with long-range spatial reasoning.

\subsection{Long-Range Dependency Modeling Beyond Self-Attention}
Self-attention excels at global context aggregation but incurs quadratic complexity and can overfit in small, noisy clinical datasets~\cite{vaswani2017attention}. Numerous efficient-attention variants alleviate the cost via low-rank projections or kernelization~\cite{wang2020linformer}. Complementarily, structured state space models (SSMs) such as S4 and subsequent Mamba-style architectures provide linear-time sequence modeling with strong long-range memory~\cite{gu2023mamba}. At the same time, recent hybrids of convolution and SSMs further expand context length without quadratic scaling~\cite{poli2023hyena}. Applications of SSMs to neuroimaging remain nascent, and our design addresses this gap by embedding a Mamba-inspired selective scanning module within the frequency-aware branch and coupling it with a cross-scale Transformer head, yielding efficient frequency–spatial integration that is both spectrum-aware and data-efficient.

\section{Method}
The RTGMFF pipeline, as illustrated in Fig.~\ref{stta}, unfolds in three distinct stages. Initially, the ROI-driven fMRI Text Generation (RFTG) deterministically summarizes region-level activation profiles together with demographic information into concise tokens. Next, a Hybrid Frequency-Spatial Encoder (HFSE) framework merges frequency-aware local cues with global spatial context: its Hierarchical Wavelet–Mamba (HWM) branch performs multi-level Haar wavelet decomposition and SelectiveScan pruning to capture fine-grained, frequency-domain structure, while its Cross-Scale Transformer Encoder (CSTE) attends across patch embeddings to inject long-range spatial dependencies and then fuses the two streams into a unified visual embedding. Finally, the Adaptive Semantic Alignment Module (ASAM) projects the ROI tokens into the visual space and optimizes a regularized alignment loss, yielding a coherent multimodal representation ready for downstream diagnosis.

\subsection{ROI-driven fMRI Text Generation (RFTG)}\label{AA}
Historically, imaging report generation has followed two primary approaches. The first involves structured reporting systems that use predefined templates filled with specific imaging features. The second approach utilizes neural encoder-decoder models, which are trained to convert visual features directly into free-form text. These models can be enhanced with knowledge graphs or retrieval-based modules. Recently, there has been interest in using general-purpose large language models (LLMs) in this field, though their adoption remains limited. While these traditional methods are intuitive, they come with several drawbacks. The two-step process can lead to random formatting errors, incur unnecessary computational costs, and raise privacy concerns when used in clinical settings. To address these issues, we propose a rule-based generator that operates deterministically on region-level activation and connectivity statistics. This approach ensures compatibility with subsequent modules and aligns with clinician-oriented reporting needs.

\textbf{ROI Statistics.}  For every subject, we begin by spatially averaging the pre-processed BOLD time-series within each of the 116 anatomical regions specified by the AAL-116 atlas. This operation yields a percentage signal change value \( \Delta\text{BOLD}_{i} \) for region \( i \), and the resulting collection 
\( \bm{v} = [\Delta\text{BOLD}_{1}, \ldots, \Delta\text{BOLD}_{116}] \) serves as a compact profile of whole-brain activity.

\textbf{Discretisation via Nested CV and Task-level Objective.}
We discretise the continuous amplitude profile \( \bm{v} \) into three ordinal bins with two thresholds \( \bm{\tau}_1 < \bm{\tau}_2 \):
{\footnotesize
\begin{align}
\mathbf{s}_i &=
\begin{cases}
    \mathbf{strong}   & \text{if } |\bm{v}_i| \ge \bm{\tau}_2, \\
    \mathbf{moderate} & \text{if } \bm{\tau}_1 \le |\bm{v}_i| < \bm{\tau}_2, \\
    \mathbf{weak}     & \text{if } |\bm{v}_i| < \bm{\tau}_1,
\end{cases}
\quad
\mathbf{p}_i =
\begin{cases}
    \uparrow   & \text{if } \bm{v}_i \ge 0, \\
    \downarrow & \text{if } \bm{v}_i < 0.
\end{cases}
\label{eq:roi_triplet}
\end{align}
}

\noindent
Each ROI is therefore represented by a triplet $\langle ROI_i,s_i,p_i\rangle$.
Crucially, we do not assume any ROI-level ground-truth labels.
Instead, the thresholds $(\tau_1,\tau_2)$ are selected by nested cross-validation to maximise the end-task validation classification accuracy.

For each outer fold, reserved for final testing, we conduct inner $K$-fold validation. We use Optuna's TPE to sample candidate pairs $(\tau_1,\tau_2)$ with the constraint $\tau_2 \ge \tau_1 + \delta$, where $\delta$ is set to 0.02. The process involves three steps: first, discretizing \( \bm{v} \) into the set $\{(s_i,p_i)\}_{i=1}^{116}$ using Eq.~\eqref{eq:roi_triplet} on the inner training split; second, serializing these triplets into tokens and training the downstream classifier along with the visual branch on the inner training split; and third, calculating the classification accuracy on the inner validation split. The classification accuracies from the inner validation are averaged across the $K$ folds, and these averages serve as the objective for evaluating each $(\tau_1,\tau_2)$. The optimal pair from the inner loop is then fixed and applied to the outer test fold, maintaining strict train-test separation and ensuring reproducibility.

\begin{figure*}[t]
\centering
\includegraphics[width=0.9\textwidth]{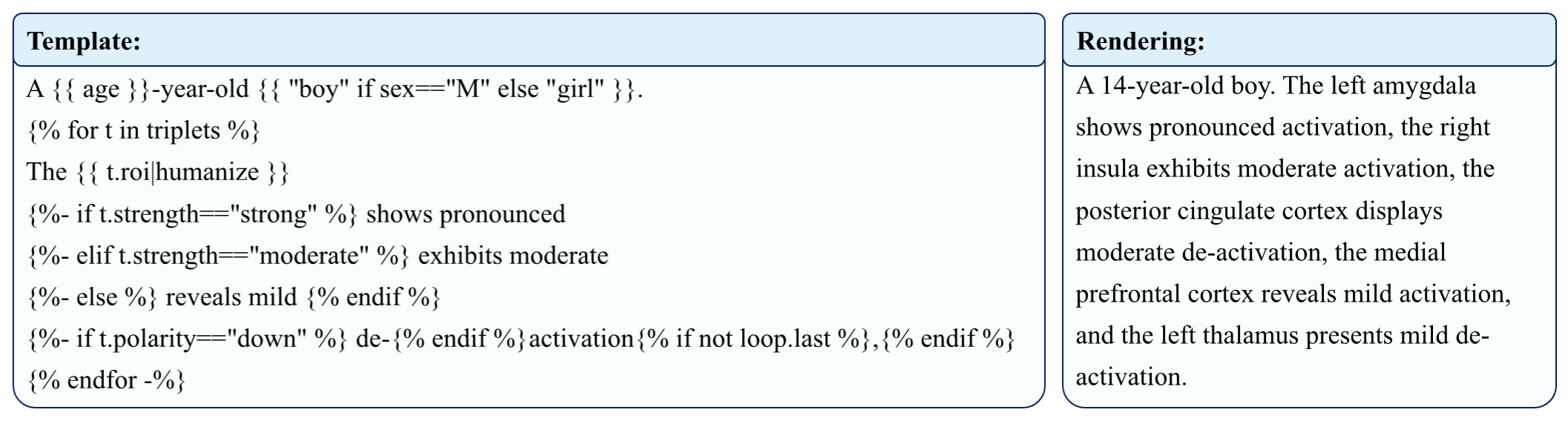}
\caption{Left, a deterministic Jinja2 template that converts subject demographics and ROI activation triplets into radiology-style prose; right, the resulting report sentence for a 14-year-old boy showing varied regional activations and de-activations.}
\label{template}
\end{figure*}

\textbf{Demographic Conditioning.} Demographic context remains clinically informative and is thus incorporated as an independent conditioning signal.
We encode chronological age and biological gender as a three-element vector:
\begin{equation}
\bm{d} = [\mathbf{age}_{\mathbf{norm}},\; \mathbf{gender}_{\mathbf{male}},\; \mathbf{gender}_{\mathbf{female}}],
\end{equation}
where the first entry normalizes age to a z-score unit scale, and the two indicator variables form a one-hot representation of gender. This vector is included with each mini-batch as a global attribute, ensuring that population context is consistently retained during training.

During feature encoding, we apply feature-wise linear modulation (FiLM) to intermediate feature maps. A two-layer multilayer perceptron takes $\bm{d}$ as input and produces a pair of scale and shift coefficients, $\gamma(\bm{d})$ and $\beta(\bm{d})$. 
Given a node representation $\bm{h} \in \mathbb{R}^H$ of hidden dimension~$H$, the modulated feature is computed as:
\begin{equation}
\mathbf{FiLM}(\bm{h};\, \bm{d}) = \gamma(\bm{d}) \odot \bm{h} + \beta(\bm{d}),
\end{equation}
where $\odot$ denotes element-wise multiplication.

\textbf{Clinician-oriented Report.}  Although the computational pipeline operates entirely on JSON-encoded
triplets, many clinical sites still require a concise narrative that
can be copied into the radiology information system.  
To meet this need, we provide an optional post-processing step based on a light-weight Jinja2 template.  
The template takes each subject’s age, gender, and the ordered list of
triplets, then renders a single English sentence per ROI, as shown in Fig.~\ref{template}. Because the template engine is rule-based and deterministic, it
introduces no additional learnable parameters, does not affect
diagnostic performance, and can be disabled entirely in settings where human-readable text is unnecessary.

\subsection{Hybrid Frequency-Spatial Encoder (HFSE)}

Effectively extracting and integrating multi-scale feature information from fMRI images is crucial for accurate analysis. Traditional methods based on CNN and Transformer often fail to utilize the potential of frequency domain information and do not fully capture the comprehensive features of fMRI data. While the Mamba structure has demonstrated excellent performance in modeling long-range dependencies in the spatial domain, its application in the frequency domain remains unexplored. To address these limitations, the HFSE framework introduces an innovative approach that combines frequency domain modeling capabilities with Mamba's ability to capture spatial long-range dependencies. This novel method offers a more comprehensive solution for fMRI feature extraction, effectively bridging the gap between spatial and frequency domain information.

\textbf{Hierarchical Wavelet-Mamba (HWM).} To extract local embeddings from an input fMRI image \( \bm{X} \in \mathbb{R}^{H \times W \times 3} \), we apply a $N$-level 2D Haar DWT recursively on the LL subband:
{\small
\begin{equation}
\bm{X}^{(0)} = \bm{X}, \quad 
\bm{X}^{(s)} = \mathbf{DWT}_{\textnormal{Haar}}\left( \bm{X}^{(s-1)} \right), \quad 
s = 1, \ldots, N
\end{equation}
}

\noindent
where each level yields four subbands and reduces the spatial resolution by a factor of 2.  
Thus,
\(
\bm{X}^{(\mathbf{s})} \in \mathbb{R}^{H/2^{\mathbf{s}} \times W/2^{\mathbf{s}} \times 4C}.
\)  
Features extracted from all decomposition levels are concatenated and flattened into a token sequence:

\begin{equation}
\bm{F}_{\text{seq}} = \mathbf{Flatten} \left( \mathop{\oplus}\nolimits_{s=1}^{N} \bm{F}_{s} \right) \in \mathbb{R}^{\bm{L} \times \bm{D}}.
\end{equation}
Next, the sequence is divided into four sub-sequences \( \{ \bm{T}_i \}_{i=1}^4 \) through a learnable linear projection. Each sub-sequence undergoes token pruning using a Mamba-inspired SelectiveScan module \cite{gu2023mamba} and semantic refinement via an FFN block:
\begin{equation}
\bm{F}_{\text{final}} = \mathbf{ConvFFN} \left( \mathbf{SelectiveScan}(\bm{T}_i) \right) \in \mathbb{R}^{\bm{L}/\bm{D}}.
\end{equation}

\textbf{Cross-Scale Transformer Encoder (CSTE).} To effectively integrate global and local representations, the input \( \bm{X} \) is first divided into \( p \times p \) patch embeddings, represented as \( \bm{E} \in \mathbb{R}^{(H/p)(W/p) \times D_p} \). These embeddings are normalized and downsampled using a stride-2 convolution to produce the query sequence \( \bm{Q}' \in \mathbb{R}^{L_q \times D} \), which serves as the basis for global representation.

Simultaneously, the multi-scale feature \( \bm{F}_{\text{final}} \) is processed to capture local context. This involves spatial smoothing and channel compression through a sequence of cascaded convolutional operations, defined as:

{\small
\begin{equation}
\bm{F}_{\text{local}} = \boldsymbol{\mathrm{AvgPool}}\left( \boldsymbol{\mathrm{ReLU}}\left( \boldsymbol{\mathrm{BN}}\left( \boldsymbol{\mathrm{Conv}}_{3\times3}\left( \boldsymbol{\mathrm{Conv}}_{1\times1}\left( \bm{F}_{\text{final}} \right) \right) \right) \right) \right).
\end{equation}
}

From this locally processed feature, key-value pairs are generated via linear projection, where \( \bm{K} = \bm{W}_k \bm{F}_{\text{local}} \) and \( \bm{V} = \bm{W}_v \bm{F}_{\text{local}} \). Cross-scale attention is then computed between the query sequence \( \bm{Q}' \) and the local features, as follows:

{\small
\begin{equation}
\bm{A} = \mathbf{Softmax} \left( \frac{\bm{Q}' \bm{K}^\top}{\sqrt{\bm{D}}} \right) \bm{V} \in \mathbb{R}^{\bm{L}_q \times \bm{D}}.
\end{equation}
}

The resulting attention output \( \bm{A} \) is upsampled and fused with the patch embeddings \( \bm{E} \) through element-wise addition. This combined representation is subsequently refined using a 4-layer Vision Transformer \cite{dosovitskiy2020image}, culminating in the final global representation \( \bm{Z} \in \mathbb{R}^{D} \).

\subsection{Adaptive Semantic Alignment Module (ASAM)}
The ROI token sequence produced by RFTG, denoted \( \bm{T}_{\text{input}} \), is first converted into contextual embeddings \( \bm{T}_{\text{emb}} \in \mathbb{R}^{D_t} \) using a pretrained BioBERT model. When a narrative report is required for clinicians, it is rendered from the same tokens via a deterministic template but is not used during model training.  
To align visual and textual features, we project both modalities into a shared latent space and compute a cosine‐similarity‐based alignment score:
{\small
\begin{equation}
\bm{\hat{y}} = \mathbf{MLP}\!\left( \cos\!\left( \bm{Z}\bm{W}_z,\; \bm{T}_{\text{emb}}\bm{W}_t \right) \right),
\end{equation}
}

\noindent
where \( \bm{W}_z \in \mathbb{R}^{D \times D_a} \) and \( \bm{W}_t \in \mathbb{R}^{D_t \times D_a} \) are learnable projection matrices that map visual and textual embeddings into the common space \( \mathbb{R}^{D_a} \).

\textbf{Model Optimization.} We optimize the proposed framework by minimizing multiple objectives: a task loss, an alignment loss, and regularization constraints. For the task loss, we utilize a cross-entropy loss \( \mathcal{L}_\text{cls} \)  for classification.

To address the modality gap, we incorporate an alignment loss based on cosine similarity:
\begin{equation}
\mathcal{L}_{\text{align}} = 1 - \frac{1}{\bm{B}} \sum_{i=1}^{\bm{B}} \frac{\bm{Z}_i \bm{W}_z \cdot \bm{T}_i \bm{W}_t}{\|\bm{Z}_i \bm{W}_z\| \, \|\bm{T}_i \bm{W}_t\|},
\end{equation}
where \( \bm{W}_z \in \mathbb{R}^{D \times D_a} \) and \( \bm{W}_t \in \mathbb{R}^{D_t \times D_a} \) are learnable projection matrices.

Finally, regularization constraints are applied to promote feature dispersion and maintain a balance between modalities:
\begin{equation}
\mathcal{L}_{\text{reg}} = \| \bm{W}_z^\top \bm{W}_z - \bm{W}_t^\top \bm{W}_t \|_F^2 .
\end{equation}

The total loss function is formulated as follows:
\begin{equation}
\mathcal{L}_{\text{total}} = \mathcal{L}_{\text{cls}} + \bm{\alpha} \mathcal{L}_{\text{align}} + \bm{\beta} \mathcal{L}_{\text{reg}},
\end{equation}
where \( \boldsymbol{\alpha} \) and \( \boldsymbol{\beta} \) are the loss weights. 

\begin{table*}[!t]
\caption{Performance Comparison on ADHD-200 and ABIDE datasets.}
\centering
\fontsize{8pt}{11pt}\selectfont
\setlength{\tabcolsep}{7pt}
\renewcommand{\arraystretch}{0.8}
\begin{tabular}{
>{\raggedright\arraybackslash}p{4.0cm}
>{\centering\arraybackslash}p{1.2cm}
>{\centering\arraybackslash}p{1.2cm}
>{\centering\arraybackslash}p{1.2cm}
>{\centering\arraybackslash}p{1.2cm}
>{\centering\arraybackslash}p{1.2cm}
>{\centering\arraybackslash}p{1.2cm}
>{\centering\arraybackslash}p{1.2cm}
>{\centering\arraybackslash}p{1.2cm}
}
\toprule
\textbf{Method} & \multicolumn{4}{c}{ADHD-200 (ADHD)\cite{bellec2017neuro}} & \multicolumn{4}{c}{ABIDE (ASD)\cite{di2014autism}} \\
\cmidrule(lr){2-5} \cmidrule(lr){6-9}
 & ACC (\%) & SEN (\%) & SPE (\%) & AUC (\%) & ACC (\%) & SEN (\%) & SPE (\%) & AUC (\%) \\
\midrule
\multicolumn{9}{l}{\textbf{Universal Method}} \\
\midrule
MLP\cite{hornik1989multilayer}         & 60.8$\pm$5.5 & 64.3$\pm$4.1 & 68.6$\pm$5.3 & 66.4$\pm$3.4 & 62.9$\pm$4.3 & 64.8$\pm$2.7 & 61.4$\pm$3.1 & 63.1$\pm$2.1 \\
ResNet-50\cite{he2016deep}             & 68.5$\pm$3.4 & 57.7$\pm$4.8 & 78.1$\pm$3.6 & 67.9$\pm$3.0 & 72.2$\pm$2.9 & 68.0$\pm$3.3 & 75.6$\pm$2.6 & 71.8$\pm$2.1 \\
VGG16\cite{simonyan2014very}           & 64.6$\pm$3.8 & 59.7$\pm$5.3 & 70.3$\pm$3.1 & 65.0$\pm$3.1 & 71.3$\pm$3.2 & 73.0$\pm$4.2 & 69.1$\pm$3.9 & 71.0$\pm$2.9 \\
Swin Transformer\cite{liu2021swin}     & 73.2$\pm$2.7 & 70.3$\pm$3.6 & 75.1$\pm$2.5 & 72.7$\pm$2.2 & 76.0$\pm$1.8 & 74.2$\pm$2.1 & 77.5$\pm$1.7 & 75.8$\pm$1.4 \\
\midrule
\multicolumn{9}{l}{\textbf{Task-Specific Method}} \\
\midrule
BrainNetCNN (Neuro'17)\cite{kawahara2017brainnetcnn} & 68.0$\pm$3.3 & 65.8$\pm$8.3 & 70.4$\pm$5.3 & 68.1$\pm$4.9 & 67.3$\pm$2.6 & 63.3$\pm$9.4 & 70.5$\pm$8.7 & 66.9$\pm$6.4 \\
BrainGNN (MIA'21)\cite{li2021braingnn}               & 64.7$\pm$3.8 & 67.9$\pm$3.5 & 62.7$\pm$4.1 & 65.3$\pm$2.7 & 69.3$\pm$3.9 & 66.8$\pm$3.3 & 72.7$\pm$3.4 & 69.8$\pm$2.4 \\
BrainGB (TMI'22)\cite{cui2022braingb}                & 65.8$\pm$2.6 & 61.3$\pm$4.9 & 70.3$\pm$3.8 & 65.8$\pm$3.1 & 63.2$\pm$2.0 & 63.8$\pm$8.1 & 60.1$\pm$6.9 & 62.0$\pm$5.3 \\
BNT (NeurIPS'22)\cite{kan2022brain}                  & 72.8$\pm$2.9 & 70.9$\pm$4.3 & 73.9$\pm$3.6 & 72.4$\pm$2.8 & 75.9$\pm$1.6 & 72.9$\pm$5.3 & 69.8$\pm$6.6 & 71.3$\pm$4.2 \\
A-GCL (MIA'23)\cite{zhang2023gcl}                    & \textcolor{blue}{77.8$\pm$4.4} & \textcolor{blue}{76.4$\pm$4.7} & 79.4$\pm$5.3 & \textcolor{blue}{77.9$\pm$3.5} & 82.9$\pm$2.1 & 82.4$\pm$2.6 & \textcolor{blue}{83.7$\pm$1.9} & \textcolor{blue}{83.1$\pm$1.6} \\
Com-BrainTF (MICCAI'23)\cite{bannadabhavi2023community} & 74.2$\pm$3.1 & 74.7$\pm$4.6 & 73.3$\pm$3.9 & 74.0$\pm$3.0 & 81.3$\pm$2.6 & 79.9$\pm$5.9 & 75.2$\pm$4.3 & 77.6$\pm$3.7 \\
MCPATS (JBHI'24)\cite{jiang2024multi}               & 74.0$\pm$3.3 & 64.3$\pm$4.7 & \textcolor{blue}{80.4$\pm$2.9} & 72.3$\pm$2.8 & 76.0$\pm$2.6 & 73.2$\pm$3.6 & 78.7$\pm$2.9 & 76.0$\pm$2.3 \\
GBT (MICCAI'24)\cite{peng2024gbt}                    & 70.4$\pm$3.6 & 68.1$\pm$4.3 & 71.5$\pm$3.9 & 69.8$\pm$2.9 & 78.0$\pm$6.6 & 79.5$\pm$9.5 & 77.3$\pm$4.0 & 78.4$\pm$5.2 \\
SCDA (PR'25)\cite{fang2025source}                    & 57.8$\pm$3.9 & 55.2$\pm$9.3 & 59.2$\pm$9.9 & 57.2$\pm$6.8 & 67.5$\pm$0.9 & 68.4$\pm$8.0 & 66.8$\pm$6.6 & 67.6$\pm$5.2 \\
KMGCN (MIA'25)\cite{zeng2025knowledge}               & 75.2$\pm$2.6 & 72.8$\pm$3.9 & 77.0$\pm$2.6 & 74.9$\pm$2.3 & \textcolor{blue}{84.7$\pm$1.3} & \textcolor{blue}{83.6$\pm$1.7} & 81.3$\pm$1.4 & 82.4$\pm$1.1 \\
\midrule
\textbf{RTGMFF (Ours)} & \textcolor{red}{80.7$\pm$2.5} & \textcolor{red}{79.5$\pm$3.0} & \textcolor{red}{81.3$\pm$2.8} & \textcolor{red}{80.4$\pm$2.1} & \textcolor{red}{86.4$\pm$1.9} & \textcolor{red}{84.5$\pm$2.7} & \textcolor{red}{87.5$\pm$2.3} & \textcolor{red}{86.0$\pm$1.8} \\
\bottomrule
\end{tabular}
\label{tab1}
\end{table*}

\begin{figure*}[t]
\centering
\includegraphics[width=0.95\textwidth]{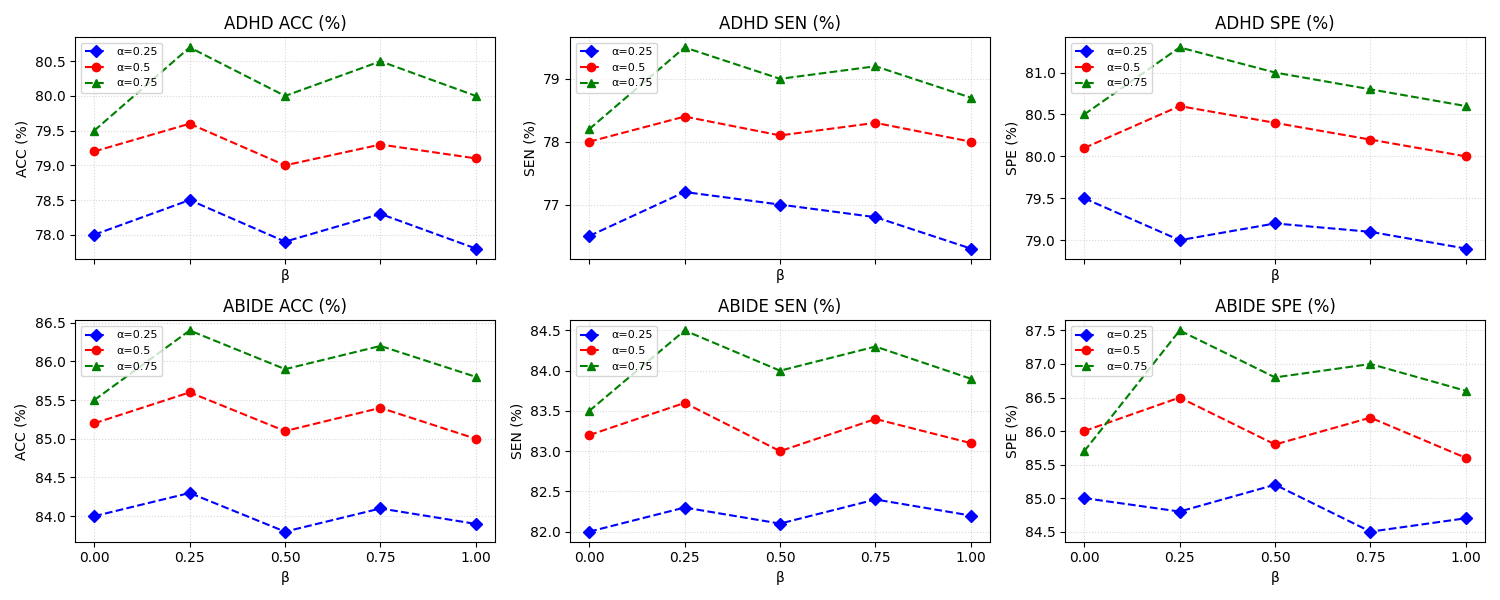}
\caption{Plots of model performance on test set versus settings of hyperparameters $\alpha$ and $\beta$.}
\label{hyper}
\end{figure*}
\begin{figure}[t]
\centering
\includegraphics[width=\columnwidth]{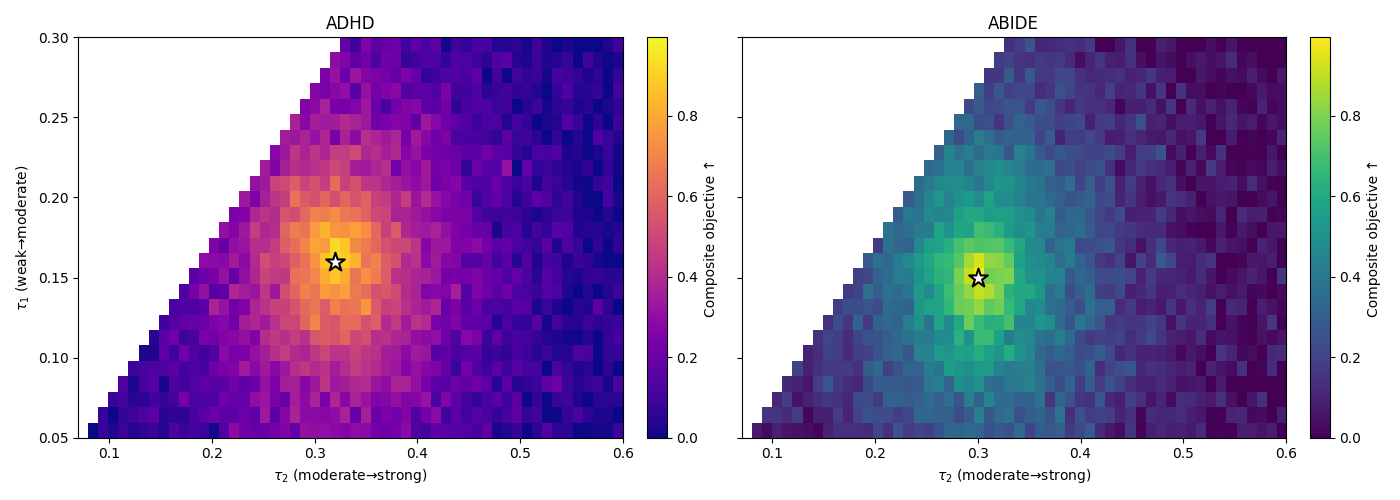}
\caption{Heatmap of macro ROI-F1 over $\tau_{1}$ (y-axis) and $\tau_{2}$ (x-axis). The white star marks the optimal $(0.15,\,0.30)$; the blank triangle is invalid due to $\tau_{2} \le \tau_{1} + 0.02$.}
\label{bayes}
\end{figure}
\begin{table*}[t]
\caption{Effect of HWM, CSTE, and ASAM on ADHD-200 and ABIDE datasets.}
\centering
\fontsize{9pt}{12pt}\selectfont
\setlength{\tabcolsep}{7pt}
\begin{tabular}{ccc cccc cccc}
\toprule
\multicolumn{3}{c}{Module} & \multicolumn{4}{c}{ADHD-200 (ADHD)\cite{bellec2017neuro}} & \multicolumn{4}{c}{ABIDE (ASD)\cite{di2014autism}} \\
\cmidrule(lr){4-7} \cmidrule(lr){8-11}
HWM & CSTE & ASAM & ACC (\%) & SEN (\%) & SPE (\%) & AUC (\%) & ACC (\%) & SEN (\%) & SPE (\%) & AUC (\%) \\
\midrule
\checkmark & \ding{55} & \ding{55} & 76.8$\pm$2.8 & 75.5$\pm$3.2 & 77.3$\pm$2.9 & 76.4$\pm$2.2 & 82.5$\pm$2.1 & 82.7$\pm$2.8 & 83.9$\pm$2.5 & 83.3$\pm$1.9 \\
\ding{55} & \checkmark & \ding{55} & 74.2$\pm$3.0 & 72.9$\pm$3.5 & 75.1$\pm$3.3 & 74.0$\pm$2.4 & 81.0$\pm$2.3 & 78.8$\pm$3.0 & 82.6$\pm$2.7 & 80.7$\pm$2.0 \\
\checkmark & \checkmark & \ding{55} & \textcolor{blue}{79.5$\pm$2.5} & \textcolor{blue}{78.1$\pm$2.9} & \textcolor{blue}{80.3$\pm$2.7} & \textcolor{blue}{79.2$\pm$2.0} & \textcolor{blue}{84.9$\pm$1.8} & \textcolor{blue}{83.2$\pm$2.5} & \textcolor{blue}{86.1$\pm$2.2} & \textcolor{blue}{84.6$\pm$1.7} \\
\checkmark & \ding{55} & \checkmark & 78.9$\pm$2.6 & 77.5$\pm$3.1 & 79.8$\pm$2.8 & 78.6$\pm$2.1 & 84.3$\pm$2.0 & 82.6$\pm$2.7 & 85.7$\pm$2.3 & 84.2$\pm$1.8 \\
\ding{55} & \checkmark & \checkmark & 76.3$\pm$2.9 & 74.8$\pm$3.4 & 77.2$\pm$3.2 & 76.0$\pm$2.3 & 82.9$\pm$2.2 & 80.9$\pm$2.9 & 84.3$\pm$2.6 & 82.6$\pm$1.9 \\
\checkmark & \checkmark & \checkmark & \textcolor{red}{80.7$\pm$2.5} & \textcolor{red}{79.5$\pm$3.0} & \textcolor{red}{81.3$\pm$2.8} & \textcolor{red}{80.4$\pm$2.1} & \textcolor{red}{86.4$\pm$1.9} & \textcolor{red}{84.5$\pm$2.7} & \textcolor{red}{87.5$\pm$2.3} & \textcolor{red}{86.0$\pm$1.8} \\
\bottomrule
\end{tabular}
\label{tab2}
\end{table*}
\section{Experiments and Discussion}
\subsection{Datasets and Implementation}
\textbf{Datasets.} We assess our proposed method using two publicly available fMRI datasets: ADHD-200 \cite{bellec2017neuro} and ABIDE \cite{di2014autism}. The ADHD-200 dataset is used for diagnosing Attention Deficit Hyperactivity Disorder (ADHD), while the ABIDE dataset focuses on classifying Autism Spectrum Disorder (ASD).

\textbf{ADHD-200 Dataset.} The ADHD-200 dataset provides fMRI data and related non-imaging information from eight international sites. Following rigorous quality control procedures, we selected 776 subjects, including both individuals with ADHD and typically developing controls.

\textbf{ABIDE Dataset.} The ABIDE dataset compiles neuroimaging and non-imaging data from 20 international sites. To ensure a fair comparison with previous state-of-the-art methods, we selected a cohort of 871 subjects from ABIDE, consisting of 468 Healthy Controls (HC) and 403 ASD patients.

\textbf{Implementation Details.} All experiments were conducted using PyTorch 2.1 on a single NVIDIA RTX 4090 with CUDA 12.4. The network was trained using the AdamW optimizer, starting with a learning rate of $1\times10^{-4}$ for the backbone and a rate five times higher for the task head, alongside a weight decay of $1\times10^{-4}$. A cosine-annealing schedule with linear warm-up was applied to control the learning rate. The backbone remained frozen for the initial five epochs as part of a linear-probing phase, after which the training proceeded end-to-end. Each mini-batch comprised eight four-dimensional fMRI volumes. Training was limited to a maximum of 120 epochs, employing early stopping with a validation-loss patience of ten epochs. 

To avoid information leakage related to imaging sites, a leave-one-site-out cross-validation protocol was used: in each fold, subjects from one imaging site formed the test set, while the remaining sites were divided 90\%/10\% into training and validation subsets. Performance is reported as the mean $\pm$ standard deviation across all folds. To ensure full reproducibility, all random seeds were fixed to 42.

\textbf{Data Representation.} Each subject's 4D fMRI data ($x{\times}y{\times}z{\times}t$) is transformed into a 2D three-channel map $X\in\mathbb{R}^{H\times W\times 3}$ through the following process. Initially, the brain is segmented using the AAL-116 atlas, and three statistical measures, ALFF, fALFF, and ReHo, are calculated from the band-limited BOLD time series (0.01–0.08 Hz) for each region of interest. These measures are then mapped to a fixed 2D atlas layout, with the left and right hemispheres arranged side by side, forming three distinct channels: $\{\text{ALFF},\text{fALFF},\text{ReHo}\}$. This 2D representation effectively maintains regional contrasts, reduces the computational demand associated with 3D processing, and facilitates frequency-aware modeling in subsequent analyses.

\begin{figure*}[t]
  \centering
  \includegraphics[width=0.95\textwidth]{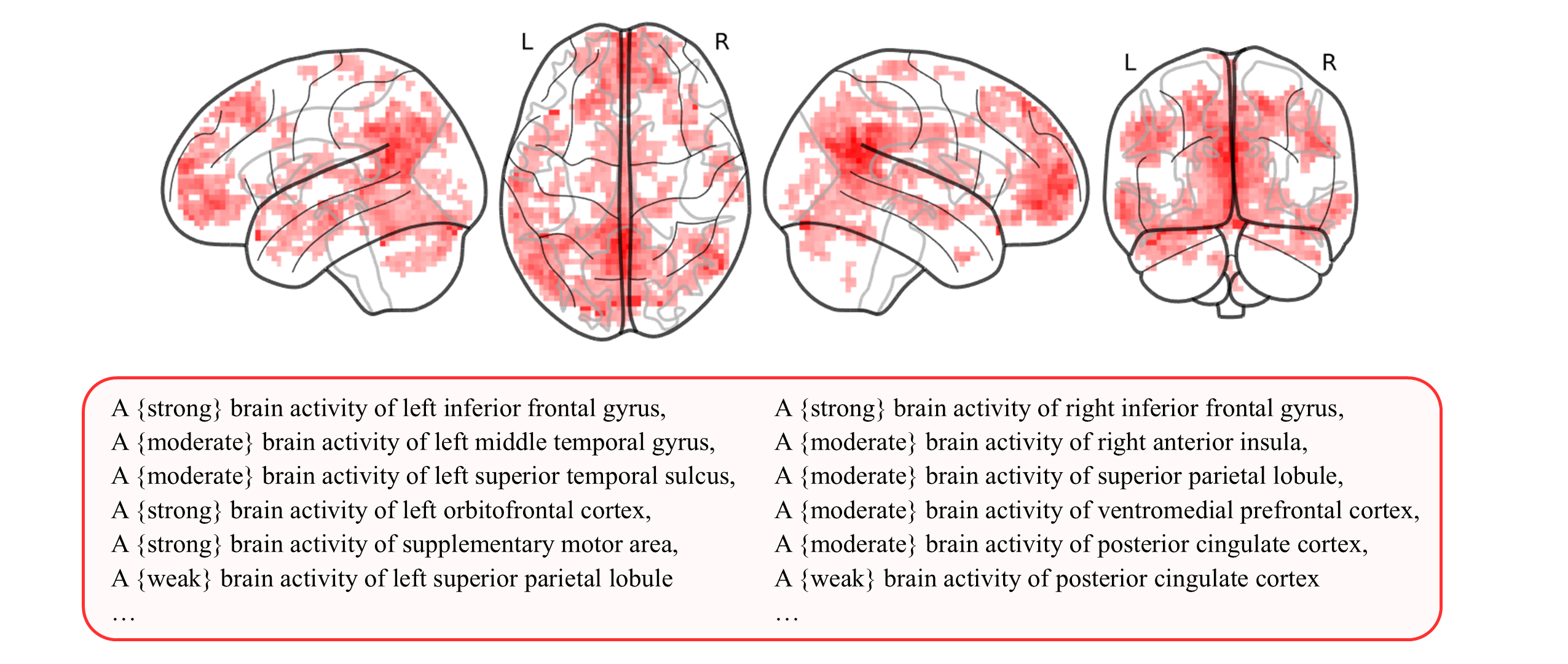} 
  \caption{Qualitative results of ROI-driven fMRI text generation.
  (\textbf{Top}) Cortical surface maps with three activation levels
  (light, medium, and dark red denote weak, moderate, and strong activity).
  (\textbf{Bottom}) Automatically generated clauses for the left (L) and
  right (R) hemispheres, illustrating the fidelity and interpretability
  of the proposed RFTG module.}
  \label{fig:rftg_vis}
\end{figure*}
\subsection{Experimental Results}
\textbf{Comparisons.}
We assessed the effectiveness of our proposed method, RTGMFF, by comparing it with a wide range of existing models across various architectures. These included CNN-based approaches like MLP \cite{hornik1989multilayer}, ResNet-50 \cite{he2016deep}, VGG16 \cite{simonyan2014very}, and BrainNetCNN \cite{kawahara2017brainnetcnn}. GNN-based methods such as BrainGNN \cite{li2021braingnn}, BrainGB \cite{cui2022braingb}, A-GCL \cite{zhang2023gcl}, and KMGCN \cite{zeng2025knowledge}. Transformer-based approaches like Swin Transformer \cite{liu2021swin}, BNT \cite{kan2022brain}, Com-BrainTF \cite{bannadabhavi2023community}, MCPATS \cite{jiang2024multi}, and GBT \cite{peng2024gbt}, as well as the unsupervised domain adaptation method SCDA \cite{fang2025source}. For evaluation, we used standard classification metrics, including Accuracy (ACC), Sensitivity (SEN), Specificity (SPE), and Area Under the Curve (AUC), where higher values indicate better performance.

Table \ref{tab1} presents a quantitative comparison of RTGMFF’s performance against these methods. Simpler models like MLP and ResNet-50 showed inferior performance due to their basic structures, whereas RTGMFF demonstrated superior results compared to other state-of-the-art methods. Notably, RTGMFF outperformed A-GCL by 2.9\% on the ADHD-200 dataset and 3.5\% on the ABIDE dataset in terms of Accuracy. Similarly, it surpassed KMGCN by 5.5\% on ADHD-200 and 1.7\% on ABIDE.

Moreover, RTGMFF exhibited significant improvements in Sensitivity, achieving 79.5\% for ADHD-200 and 84.5\% for ABIDE. The Specificity metric also improved, reaching 81.3\% for ADHD-200 and 87.5\% for ABIDE. In terms of the AUC metric, which reflects the model's overall discriminative ability, RTGMFF achieved scores of 80.4\% on the ADHD-200 dataset and 86.0\% on the ABIDE dataset, further affirming its robustness and superior classification performance.

\textbf{Ablation Study.} As shown in Table~\ref{tab2}, our ablation study reveals that the removal of the HWM module results in a significant performance decline of over 4\% across all metrics on both datasets. This underscores the module's essential role in enhancing model performance. Building on the HWM module, the addition of the CSTE module leads to a noticeable improvement of over 3\% in accuracy metrics on both datasets. The AUC metric also improves, reaching 79.2\% on the ADHD-200 dataset and 84.6\% on the ABIDE dataset, confirming the module's effectiveness in integrating local and global information. Further enhancement is observed with the integration of the ASAM module, which provides an additional performance boost of over 2\%. The AUC metric increases further, reaching 80.4\% on the ADHD-200 dataset and 86.0\% on the ABIDE dataset, validating the module's effectiveness. These findings from the ablation study collectively confirm the value of each module, highlighting their significant contributions to elevating model performance.

\subsection{Hyper-Parameter Sensitivity Analysis}
\textbf{Loss Function.} In our framework, we introduce two key hyperparameters, $\alpha$ and $\beta$, to balance alignment and redundancy constraints during cross-scale fusion. $\alpha$ controls the alignment strength between modalities, while $\beta$ regulates regularization intensity to reduce feature redundancy. We evaluate their effects through a grid search with $\alpha \in \{0.2, 0.5, 0.8\}$ and $\beta \in \{0.1, 0.2, 0.3\}$ on the ADHD-200 and ABIDE datasets.

Fig.~\ref{hyper} shows how variations in $\alpha$ and $\beta$ affect Accuracy (ACC), Sensitivity (SEN), and Specificity (SPE). For $\alpha$, a low value of 0.2 causes weak alignment and suboptimal fusion, while increasing to 0.5 improves performance, and $\alpha = 0.8$ yields the best results. Further increases, however, slightly reduce performance due to over-alignment.

Similarly, $\beta = 0.1$ fails to suppress redundancy, whereas $\beta = 0.3$ over-regularizes and limits feature learning. The optimal balance is found at $\beta = 0.2$, enhancing discriminability and generalization. Thus, $\alpha = 0.8$ and $\beta = 0.2$ are adopted for all subsequent experiments.

\textbf{ROI Discretization Thresholds.} As shown in Fig.~\ref{bayes}, we applied nested cross-validation with Optuna’s TPE sampler across 100 trials to optimize the macro-averaged ROI-F1 score. The search covered $\tau_{1}$ from 0.05 to 0.45 and $\tau_{2}$ from $\tau_{1}+0.02$ to 0.60. Five outer folds were used for final evaluation, and three inner folds assessed each $(\tau_{1}, \tau_{2})$ pair. For each subject, the 116-dimensional activation vector was discretized, and macro F1 was computed and averaged across inner folds.

The optimal thresholds were consistently around $(\tau_{1}, \tau_{2}) = (0.15, 0.30)$ for both ABIDE and ADHD-200, with ADHD-200 showing a slight peak at $(0.16, 0.32)$. To maintain strict train-test separation and reproducibility, we fixed $(\tau_{1}, \tau_{2}) = (0.15, 0.30)$ for all test folds.

\subsection{Qualitative Visualization of RFTG}
Fig.~\ref{fig:rftg_vis} showcases two representative fMRI samples. The surface plots illustrate varying levels of activation using light, medium, and dark red tones, respectively. In the lower panels, the sentences generated by our RFTG module are presented. The precise alignment between the highlighted regions and the textual descriptions demonstrates that the RFTG module effectively translates quantitative BOLD statistics into anatomically accurate, comprehensible reports.

\section{Conclusion}
We present RTGMFF, a deterministic multimodal diagnostic framework integrating frequency-aware and global cues via HWM, CSTE, and ASAM modules. On the ADHD-200 and ABIDE benchmarks, the method demonstrates superior performance, surpassing established baselines. Future work will incorporate richer clinical data, extend to 4D spatiotemporal modeling, compare performance against lightweight LLM-based generative approaches, and explore clinical validation of the auditable text tokenization by comparing RFTG outputs against expert-drafted radiology reports. RTGMFF provides a robust foundation for reliable multimodal fMRI diagnostics.

\section{ACKNOWLEDGEMENTS}

This work was supported in part by the Fundamental Research Funds for the Provincial Universities of Zhejiang (No. GK259909299001-006), the 'Pioneer' and 'Leading Goose' R\&D Program of Zhejiang (No. 2025C04001), and the Anhui Provincial Joint Construction Key Laboratory of Intelligent Education Equipment and Technology (No. IEET202401).

\bibliographystyle{IEEEtran}
\bibliography{reference}

\end{document}